\documentclass[11pt]{article}
\pdfoutput=1
\usepackage{authblk}
\usepackage[T1]{fontenc}
\usepackage[utf8]{inputenc}
\usepackage{coling2020,microtype,times,latexsym,booktabs,mdwlist,natbib,xcolor,hyperref,todonotes}
\usepackage{float}
\definecolor{darkblue}{rgb}{0, 0, 0.5}  
\hypersetup{colorlinks=true,citecolor=darkblue, linkcolor=darkblue, urlcolor=darkblue}
\usepackage{tipa}

\title{BERTje: A Dutch BERT Model \vspace*{.5cm}}

\author[]{\textbf{Wietse de Vries}}
\author[]{\textbf{Andreas van Cranenburgh}}
\author[]{\textbf{Arianna Bisazza}\vspace*{-.2cm}}
\author[]{\\ \textbf{Tommaso Caselli}}
\author[]{\textbf{Gertjan van Noord}}
\author[]{\textbf{Malvina Nissim}}

\affil{CLCG, University of Groningen, The Netherlands
    \vspace{5px} \\
    \texttt{research@wietsedv.nl}\\
    \texttt{\{a.w.van.cranenburgh,a.bisazza,} \\
    \texttt{t.caselli,g.j.m.van.noord,m.nissim\}@rug.nl} 
} 

\date{}

\begin{document}
\maketitle

\vspace*{.7cm}
\begin{abstract}
  The transformer-based pre-trained language model BERT has helped to improve state-of-the-art performance on 
  many natural language processing (NLP) tasks.
  Using the same architecture and parameters, we developed and evaluated a monolingual Dutch BERT model called BERTje.
  Compared to the multilingual BERT model, which includes Dutch but is only based on Wikipedia text, BERTje is based on a large and diverse dataset of 2.4 billion tokens.
  BERTje consistently outperforms the equally-sized multilingual BERT model on downstream NLP tasks (part-of-speech tagging, named-entity recognition, semantic role labeling, and sentiment analysis).
  Our pre-trained Dutch BERT model is made available at \url{https://github.com/wietsedv/bertje}.
\end{abstract}

\section{Introduction}
In the field of computational linguistics there has been a major transition from the development of task-specific models built from scratch
to fine-tuning approaches based on large general-purpose language models \citep{Howard:2018,Peters:2018}.
Currently, the most commonly used pre-trained model of this type is BERT~\citep{devlin2019bert}.
This model and its derivatives are based on the transformer architecture~\citep{vaswani2017attention}.
Many state-of-the-art results on benchmark natural language processing (NLP) tasks have been improved by fine-tuned versions of BERT and BERT-derived models.

The BERT model is pre-trained with two learning objectives that force the model to learn semantic information within and between sentences~\citep{devlin2019bert}.
The masked language modeling (MLM) task forces the BERT model to embed each word based on the surrounding words.
The next sentence prediction (NSP) task, on the other hand, forces the model to learn semantic coherence between sentences.
For BERT, NSP is implemented through a binary prediction task where two sentences are either consecutive (positive instance) or the second sentence is completely random (negative instance).
It has however been shown that this method is ineffective~\citep{liu2019roberta}.
The NSP was intended to learn inter-sentence coherence, but apparently BERT actually learned topic similarity.
Indeed, if the next sentence is random, it is not just a matter of coherence: crucially, the topic is likely different.
Because of this, the authors of RoBERTa removed the NSP task from the pre-training process~\citep{liu2019roberta}.
The developers of ALBERT, instead, implemented a different solution by replacing the NSP task with a sentence order prediction (SOP) task~\citep{lan2019albert}.
In SOP, two sentences are either consecutive or swapped. This change has resulted in improved downstream task performance.

The success of BERT on NLP tasks has mostly been limited to the English language since the main BERT model is trained on English~\citep{devlin2019bert}.
For other languages, one could either train language-specific models with the same BERT architecture, or use the existing multilingual BERT model.\footnote{\url{https://github.com/google-research/bert/blob/master/multilingual.md}}
This is a model trained on all Wikipedia pages of 104 different languages, including Dutch.
However, a monolingual model may perform better at tasks in a specific language and Wikipedia is a specific domain that is not representative of general language use.
Monolingual models with the BERT architecture have been developed for Italian~\citep{polignano2019alberto},
French~\citep{le2019flaubert},
German,\footnote{\url{https://deepset.ai/german-bert}}
Finnish~\citep{virtanen2019multilingual},
and Japanese.\footnote{\url{https://github.com/cl-tohoku/bert-japanese}}
The Italian model is pre-trained on Twitter data, which may not be representative for general use of language and is only trained on the MLM objective, as the NSP task is barely applicable to tweets.
The other models are pre-trained on a combination of Wikipedia with additional data from for instance online news articles.\footnote{A monolingual Dutch model has also been made available at \url{http://textdata.nl}, but this this model was consistently significantly outperformed by multilingual BERT in our experiments.}

To demonstrate the effectiveness of using multi-genre data in a monolingual model, and to equip NLP research on Dutch with a high-performing model, we developed a Dutch BERT model which we call BERTje.\footnote{The suffix \emph{-je} is used to form diminutives in Dutch; it is also used with names in an affectionate sense. BERTje is pronounced  [\textipa{"bEr\textteshlig@}].} 
In this paper we describe the training process of BERTje and evaluate its performance by fine-tuning the model on several Dutch NLP tasks.
We compare the performance on all tasks to that achieved using multilingual BERT.

\section{Pre-training data and parameters}
To facilitate comparison and due to limited resources,
we opt to train a single Dutch BERT-based model
that is architecturally equivalent to the BERT$_\textsc{base}$ model with 12 transformer blocks~\citep{devlin2019bert}.
However, the pre-training data is of course different
and other pre-training data generation modifications were made based on later derivations of BERT.
Nevertheless, we aimed to collect a dataset of similar size and diversity as used for the English BERT model.

\subsection{Data}
For pre-training, we combined several corpora of high quality Dutch text, listed below.
The sizes in parentheses are the uncompressed text sizes after cleaning.

\begin{itemize*}
    \item Books: a collection of contemporary and historical fiction novels (4.4GB)
    \item TwNC~\citep{ordelman2007twnc}: a Multifaceted Dutch News Corpus (2.4GB)
    \item SoNaR-500~\citep{sonar500}: a multi-genre reference corpus (2.2GB)
    \item Web news: all articles of 4 Dutch news websites from January 1, 2015 to October 1, 2019 (1.6GB)
    \item Wikipedia: the October 2019 dump (1.5GB)
\end{itemize*}

Documents that originate from chats or Twitter were removed from the SoNaR corpus because of quality considerations.
We also removed the Wikipedia documents from SoNaR to avoid overlap with the full Wikipedia dump.
Finally, in order to avoid any overlap with texts that we want to use as test data, we removed all documents from SoNaR-500 that are included in the manually annotated SoNaR-1 and Lassy Small~\citep{lassy} datasets. As a result, the final pre-training dataset contains 12GB of uncompressed text which amounts to about 2.4B tokens.

Like BERT, we constructed a WordPiece vocabulary with a vocabulary size of 30K tokens.
A SentencePiece model~\citep{kudo2018sentencepiece} was created for this based on the raw pre-training dataset.
The resulting vocabulary is translated to WordPiece format for compatibility with the original BERT model.

\subsection{Pre-training procedure}
BERT was pre-trained with two objectives:
next sentence prediction (NSP) and masked language modeling (MLM).
Based on findings after the initial release of BERT, we made modifications in the pre-training data generation procedure for both tasks.

Because of the demonstrated ineffectiveness of the NSP task during pre-training, BERTje is trained with the SOP objective.
This means that the second sentence in each training example is either the next or the previous sentence. 
We also apply a different strategy for the MLM objective.
Many words are split into multiple WordPiece tokens and some suffixes of words are \textit{too easy} to predict~\citep{lan2019albert}.
Therefore, instead of randomly masking single word pieces, we mask consecutive word pieces that belong to the same word.
We masked 15\% of all tokens using this strategy.
Of these selected tokens, 80\% are replaced with a special mask token,
10\% are replaced by a completely random token,
and 10\% are left as-is.
This strategy is used to ensure that the model also accurately embeds unmasked words.

BERTje is pre-trained for 1 million iterations.
To gauge the effect of the number of iterations on the performance
of downstream tasks,
we also evaluate fine-tuning performance at the 850k iterations checkpoint.

\section{Tasks and test data}
To evaluate the effectiveness of BERTje for use on downstream tasks, the model is fine-tuned for several NLP tasks.
We use annotated data from three sources for this.

First, we use the Dutch CoNLL-2002 named-entity recognition (NER) data~\citep{tjong2002conll}.
This is a four-class BIO-encoded named-entity classification task with the following four classes: \textit{person}, \textit{organisation}, \textit{location} and \textit{miscellaneous}.
Second, we evaluate on the 16 universal part-of-speech (POS) tags in the Lassy Small treebank~\citep{lassy} part of Universal Dependencies v2.5~\citep{ud25}.
Both datasets are already split into train, development, and test sets.

Third, we evaluate on several classification tasks that originate from the SoNaR-1 corpus of written Dutch~\citep{sonar1}. We evaluate on token-level NER (6 labels), coarse POS tags (12 labels) and fine-grained POS tags (241 labels in total of which 223 are present in the training data). The fine-grained POS tags contain many labels that are used only once. In addition to NER and POS tags, we extract three other annotation types from SoNaR-1:

\begin{description*}
  \item[Semantic Role predicate-argument structures:] The semantic role label annotations in SoNaR contain predicate-argument relations. We only extract predicate and argument labels. Just the highest level labels are used, so entire subordinate clauses are considered to be a single argument and the arguments and predicates within these subordinate clauses are ignored.
  \item[Semantic Role modifiers:] Modifier phrases for semantic roles are often short and non-overlapping. The labels for this task are the modifier phrase types, regardless of the predicate they belong to.
  \item[Spatio-temporal Relations:] A subset of spatio-temporal annotations are extracted including geographical relations and verb tenses.
\end{description*}

\noindent Each of these annotations are flattened from hierarchical annotations to token-level classifications.
We use 80\% of the resulting documents for training,
10\% for validation, and 10\% for testing.
The extracted annotations are split on document level,
so there is no document overlap between splits.

Each of the previous tasks describe a low level linguistic type of information. However, we want to test BERTje on a more high-level, downstream task, such as sentiment analysis. For this, we use the 110k Dutch Book Reviews Dataset~\citep{burgh2019merits},
a balanced collection of positive and negative reviews which lends itself to a binary sentiment classification task.

\section{Results}

For each of the previously described tasks, three models are fine-tuned: multilingual BERT base, BERTje at the 850K checkpoint (BERTje$_{850k}$)
and the fully trained BERTje model (1M checkpoint).
All models are fine-tuned for four epochs on the training data for each task with the same hyperparameters.
Longer training has shown degradation of performance on some validation data and increase of performance after the fourth epoch has not been observed.

\paragraph{Named-Entity Recognition}

\begin{table}[t]
\begin{center}
  \begin{tabular}{l || c c c | c c c}
    \toprule
                   & \multicolumn{3}{c|}{CoNLL-2002} & \multicolumn{3}{c}{SoNaR-1}                                                   \\
    Model          & Train                           & Dev                       & Test           & Train & Dev   & Test           \\
    \midrule
    \citet{wu2019surprising} & - & - & \textbf{90.9}    & - & - & - \\
    \midrule
    Multilingual BERT          & 95.4                           & 81.3                     & 80.7          & 95.3 & 85.0 & 79.7          \\
    BERTje$_{850k}$ & 97.7                           & 87.7                     & 87.6          & 95.9 & 85.2 & 81.1          \\
    BERTje         & 98.0                           & 87.8                     & 88.3 & 96.8 & 86.1 & \textbf{82.1} \\
    \bottomrule
  \end{tabular}
\end{center}
   \caption{\label{table:ner}NER F1 scores according to the CoNLL-2002 evaluation script~\citep{tjong2002conll}.}
\end{table}

\autoref{table:ner} shows the span-based F1 scores of the fine-tuned models.
For both the CoNLL-2002 data as well as the SoNaR-1 data, it is clear that BERTje outperforms the multilingual BERT model.
Additionally, the BERTje model has improved after the 850K checkpoint.
Our models do not outperform the state-of-the-art test score of 90.9\% of~\citet{wu2019surprising} on the CoNLL-2002 test data.
This model is a well optimized fine-tuned large multilingual BERT model.
Based on the performance difference between multilingual BERT and BERTje, it is likely that replicating their approach with a monolingual Dutch BERT model would improve the state-of-the-art performance.

\paragraph{Part-of-Speech tagging}

\begin{table}[t]
\begin{center}
  \begin{tabular}{l || c c c | c c c | c c c}
    \toprule
                   & \multicolumn{3}{c|}{UD-LassySmall} & \multicolumn{3}{c|}{SoNaR-1 (coarse)}    & \multicolumn{3}{c}{SoNaR-1 (fine-grained)}                                                \\
    Model          & Train   & Dev  & Test  & Train & Dev  & Test &                          Train & Dev  & Test              \\
    \midrule
    Multilingual BERT & 95.1 & 92.9 & 92.5                 & 99.7  & 98.1 & 98.3             & 98.8 & 96.4 & 96.2      \\
    BERTje$_{850k}$   & 99.6 & 96.8 & \textbf{96.6}        & 99.8  & 98.6 & \textbf{98.6}    & 99.4 & 97.0 & 96.6 \\
    BERTje            & 99.6 & 96.7 & 96.3                 & 99.8  & 98.6 & 98.5             & 99.5 & 97.0 & \textbf{96.8} \\
    \bottomrule
  \end{tabular}
   \caption{\label{table:pos} Part-of-speech tagging accuracy scores for Lassy Small and SoNaR.}
\end{center}
\end{table}

\begin{table}[H]
\begin{center}
  \begin{tabular}{l || c c c | c c c | c c c}
    \toprule
                   & \multicolumn{3}{c|}{SRL Predicate-arguments} & \multicolumn{3}{c|}{SRL Modifiers} & \multicolumn{3}{c}{STR}                                                   \\
    Model             & Train & Dev & Test           & Train & Dev & Test           & Train & Dev   & Test \\
    \midrule
    Multilingual BERT & 90.8 & 79.3 & 80.4           & 77.5 & 61.8 & 62.4           & 67.9 & 63.0 & 57.3 \\
    BERTje$_{850k}$   & 96.4 & 84.0 & 85.2           & 88.5 & 66.0 & \textbf{67.3}  & 81.9 & 65.6 & 62.5 \\
    BERTje            & 96.3 & 84.3 & \textbf{85.3}  & 88.5 & 66.2 & 67.2           & 81.9 & 68.5 & \textbf{64.3} \\
    \bottomrule
  \end{tabular}
   \caption{\label{table:srl}Semantic Role Labeling (SRL) F1 scores according to the CoNLL-2002 evaluation script~\citep{tjong2002conll} and Spatio-Temporal Relation (STR) macro F1 scores.}
\end{center}
\end{table}

\begin{table}[H]
\begin{center}
  \begin{tabular}{l || c c}
    \toprule
    Model             & Train & Test \\
    \midrule
    ULMFiT, \citet{burgh2019merits} & -   & \textbf{93.8} \\
    \midrule
    Multilingual BERT & 86.5 & 89.1 \\
    BERTje$_{850k}$   & 93.8 & 92.8 \\
    BERTje            & 93.6 & 93.0 \\
    \bottomrule
  \end{tabular}
   \caption{\label{table:sent}Sentiment Analysis accuracy scores on the 110k Dutch Book Reviews Dataset.}
\end{center}
\end{table}

\autoref{table:pos} illustrates POS tagging performance of our models. BERTje does outperform multilingual BERT consistently, but the 850K checkpoint model appears to perform just as well as the fully trained BERTje model.
For all three tag sets, the difference between the 850K checkpoint and the fully pre-trained BERTje model is at most 0.3 percentage points.
This indicates that the model has already learned the relevant information before the 850K checkpoint.
This is important to acknowledge since the previously mentioned NER results shows that the model does learn new information that is relevant for named-entity recognition after this checkpoint.

For the Lassy Small dataset, BERTje outperforms the 95.98\% accuracy score achieved by UDPipe 2.0 \citep{straka2018udpipe}.
These scores are not strictly comparable, since they evaluate on UD 2.2, while we evaluate on UD 2.5; however, the differences can be assumed to be minimal.

\paragraph{Semantic Roles and Spatio-Temporal Relations}

The results in \autoref{table:srl} show that BERTje outperforms multilingual BERT for the semantic role labeling (SRL) and spatio-temporal relation (STR) based test data.
However, for these tasks the model has not really improved after the 850K checkpoint. 
For evaluation of the SRL data, the CoNLL-2002 evaluation script is used in order to take chunk overlap of multi-token expressions into account.
The results on these tasks are stand-alone since we are not aware of the existence of similar systems.

The results in \autoref{table:ner} and \autoref{table:srl} both show a similar pattern where the scores on training data are higher than the development and test results. This indicates that BERTje may be prone to overfitting just like other models. Therefore, hyper-parameter tuning for specific tasks may help to improve performance.

\paragraph{Sentiment Analysis}

\autoref{table:sent} shows the sentiment analysis accuracy scores on the 110k Dutch Book Reviews Dataset.
Without hyperparameter tuning, BERTje comes close to the 93.8\% score
that \citet{burgh2019merits} obtain with manual hyperparameter tuning of an ULMFiT model~\citep{Howard:2018}.

\section{Conclusion}
We have successfully pre-trained, fine-tuned and evaluated a Dutch BERT-based model called BERTje.
This model consistently outperforms multilingual BERT on word-level NLP tasks.
Even though multilingual BERT has been shown to perform well on Dutch NLP tasks \citep{wu2019surprising}, our results indicate that a monolingual model should be preferred.

In addition to the comparison with multilingual BERT, we see that lower level linguistic structure like part-of-speech tags appear to be learned earlier during pre-training than higher level information.
Low-level linguistic tasks do not benefit from longer pre-training after 850K epochs, but the higher-level entity recognition task does benefit from longer pre-training.
This gives an indication that higher level structures in language are only properly learned after lower level structures have been encoded.
Therefore, it is important that large pre-trained language models are trained for enough iterations to properly encode high level structures.
It has been observed that English BERT encodes higher level linguistic structures in later layers~\citep{jawahar2019does, tenney2019bert} and this may be the case for BERTje too.

In future work, the encoding of different layers of linguistic abstraction within BERTje should be explored in order to fully understand and evaluate how well BERTje has learned different types of information.
It also needs to be investigated how well BERTje performs on sentence-level tasks that require coherence information between sentences.

\section*{Acknowledgments}
We are grateful to Daniel de Kok for sharing the Wikipedia data. 
BERTje was trained with Cloud TPUs from Google's TensorFlow Research Cloud (TFRC).

\end{document}